\documentclass{aadebug}

\begin{document}

\corr{0309027}{211}

\runningheads{Daniel Sundmark et al.}{Replay Debugging of
Complex Real-Time Systems: Experiences from Two Industrial Case
Studies}

\title{
Replay Debugging of Complex Real-Time Systems:
Experiences from Two Industrial Case Studies
}

\author{
Daniel~Sundmark\addressnum{1}\comma\extranum{1},
Henrik~Thane\addressnum{1}\comma\extranum{1},
Joel~Huselius\addressnum{1}\comma\extranum{1},
Anders~Pettersson\addressnum{1}\comma\extranum{1},
Roger~Mellander\addressnum{2}, 
Ingemar~Reiyer\addressnum{2},
Mattias~Kallvi
}

\address{1}{
Department of Computer Science and Engineering,
M\"{a}lardalen University,
PO Box 883,
SE-721 23 V\"{a}ster\aa{}s,
Sweden
}

\address{2}{
ABB Robotics
Hydrov\"{a}gen 10, 
SE-721 68 V\"{a}ster\aa{}s,
Sweden
}

\extra{1}{E-mail: \{daniel.sundmark, henrik.thane, joel.huselius, anders.pettersson\}@mdh.se}

\pdfinfo{
/Title (Replay Debugging of Complex Real-Time Systems:
Experiences from Two Industrial Case Studies)
/Author (Daniel Sundmark et al.)
}

\begin{abstract}
Deterministic replay is a method for allowing complex multitasking
real-time systems to be debugged using standard interactive
debuggers. Even though several replay techniques have been proposed
for parallel, multi-tasking and real-time systems, the solutions have
so far lingered on a prototype academic level, with very little
results to show from actual state-of-the-practice commercial
applications. This paper describes a major deterministic replay
debugging case study performed on a full-scale industrial robot
control system, as well as a minor replay instrumentation case study
performed on a military aircraft radar system. In this article, we
will show that replay debugging is feasible in complex multi-million
lines of code software projects running on top of off-the-shelf
real-time operating systems. Furthermore, we will discuss how replay
debugging can be introduced in existing systems without impracticable
analysis efforts. In addition, we will present benchmarking results
from both studies, indicating that the instrumentation overhead is
acceptable and affordable.
\end{abstract}


\section{Introduction}

ABB is a world leading manufacturer of industrial robots for
industrial automation. Out of all deployed industrial robots to
date, ABB has delivered about 50 percent. Out of those 50
percent, 70 percent are deployed in the car manufacturing
industry.
SAAB Avionics is a major supplier of electronic warfare
technology on the international market. The main focus of the
company is electronic warfare systems, such as display systems,
tactical reconnaissance systems and electromagnetic technology
services. Avionics products can for example be found in the
Swedish fighter aircraft Gripen, the American F-15 and the NH-
90 helicopter.

\subsection{Contribution}

In this paper, we present results from two industrial case studies
performed in cooperation with the above companies. With these case
studies, we show that our recent research results have not merely been
academic artifacts, but contributions to a fully operational method of
debugging full-scale industrial real-time systems. In addition, we
present benchmarking results from both case studies, showing that the
overhead incorporated in the system by instrumentation is acceptable.

\subsection{Paper Outline}

The rest of this paper is organized as follows: Section~\ref{bg} will
give a background and a motivation to the case studies as well as
short descriptions of the systems studied. Section~\ref{ti} describes
the implementations of our method in the investigated systems. In
Section~\ref{bm}, instrumentation benchmarking results are presented.
Finally, Section~\ref{cs} and Section~\ref{fw} conclude the paper and
discuss future work.

\section{Background and Motivation}\label{bg}

It is no secret that testing, debugging and maintenance constitute the
largest percentage of the overall cost of an average industrial
software project. In a recent study, NIST \cite{nist2002} has shown
that more than \$59 billion/year is spent on debugging software in the
U.S.A. alone. As the average complexity of software applications
increase constantly it is now common that testing and debugging
constitute more than 80\% of the total life cycle cost
\cite{nist2002}. A known fact is also that bugs are introduced early
in the design but not detected until much later downstream in the
development cycle, typically during integration, and early customer
acceptance test. For embedded real-time software this fact makes the
situation really difficult, since most failures that are detected
during integration and early deployment are extremely difficult to
reproduce. This makes debugging of embedded real-time systems costly
since repetitive reproductions of the failure is necessary in order to
find the bug.  The lack of proper tools and methods for testing and
debugging of complex real-time systems does not help the
situation. The reason why ABB Robotics and SAAB Avionics systems were
chosen as case study subjects was based on their high level of
software- and overall technical complexity. In addition, the systems
operate in safety-critical and high availability environments, where
failures might be very costly, making system validation and
verification even more important.

\subsection{Replay Debugging}

During the mid-eighties, in an effort to address the problems with
inadequate tools for debugging of complex systems, LeBlanc and
Mellor-Crummey proposed a method of recording information during
run-time and using this information to reproduce the exact behavior of
the execution off-line \cite{leblanc1987}. This method, called
\emph{Instant Replay}, allowed otherwise unfit cyclic debugging
techniques to be used for debugging nondeterministic systems. Instant
Replay, as many of its successors
\cite{audenaert1994}\cite{choi2001}\cite{ronsse1999}\cite{tai1991}\cite{chassin2000},
was focused on debugging of non-real-time concurrent programs, thereby
concentrating mainly on the correct reproduction of synchronization
races and, in some cases, on-the-fly detection of data races. However,
some methods for debugging of complex real-time systems have also been
proposed \cite{dodd1992}\cite{tsai1990}. These methods have called for
the availability of specialized hardware and non standard
instrumentation tools in order to work satisfactorily. They have also
been mere academic prototypes and not suited for the complexity of
real world software applications.

\subsection{Real-Time System Debugging using Time Machines and
Deterministic Replay}

In our previous work on debugging of embedded real-time systems, we
have proposed a replay method for recording important events and data
during a \emph{reference execution} and to reproduce the execution in
a subsequent \emph{replay execution}
\cite{thane2000a}\cite{thane2003}\cite{huselius2002a}. As for most
replay debugging methods, if the reference execution fails, the replay
execution can be used to reproduce the failure repeatedly in a
debugger environment in order to track down the bug. Our method allows
replay of real-time applications running on top of standard commercial
real-time operating systems (RTOS) and uses standard cyclic debuggers
for sequential software. There is no need for specialized hardware or
specialized compilers for the method to work and the software based
instrumentation overhead has so far proven acceptable.  This allows
our probes to be left permanently in the system, making post-mortem
debugging of a deployed system possible while at the same time
eliminating the risk of experiencing \emph{probe effects}
\cite{gait1986} during debugging. We refer to our method as
\emph{Deterministic Replay}. The tool that is used to ``travel back in
time'' and investigate what sequence of events that led to a failure
is referred to as the \emph{Time Machine} and consists of three major
parts: The \emph{Recorder} which is the instrumentation mechanism
(similar to the black-box or flight recorder in an airplane), the
\emph{Historian} which is the post-mortem off-line analysis tool and
the \emph{Time Traveler} which is the mechanism that forces the system
to behave exactly as the reference execution off-line.

The main objective of performing the case studies was to
validate the applicability of the Time Machine method to
existing complex real-time systems. The basic questions were:

\begin{itemize}

\item{Would it be possible to reproduce the behavior of such
complex target systems?}

\item{How do we minimize the analysis and implementation
effort required in order to capture the data required
from such complex target applications?}

\item{Would the instrumentation overhead (execution time
and data bandwidth) be sufficiently low for
introduction in real-world applications?}

\end{itemize}

\subsection{ABB Robotics System Model}

To validate our ideas, we had the opportunity to work with a
robotics system that is state-of-the-art in industrial
manufacturing automation. The system is a highly complex
control system application running on top of the commercial
Wind River RTOS VxWorks. However, in order to avoid the
somewhat impossible mission of analyzing and instrumenting an
approximate of 2.5 million lines of code in an academic project,
we focused on instrumenting five central parts in the target
system:

\begin{itemize}

\item{The operating system.}

\item{The application's operating system abstraction layer.}

\item{The inter-process communication abstraction layer.}

\item{The peripheral I/O abstraction layer.}

\item{The state preserving structures of each individual task.}

\end{itemize}

All instrumentation was system wide and application-transparent except
for the state preserving structures in the tasks. A more thorough
description of the instrumentation will be given below. For an
overview, see Figure~\ref{instrumentation}.

Out of an approximate total of 70 tasks, we choose to record the state
(internal static variables) of the three most frequently running
tasks. Thus, we limited the instrumentation efforts for parts of the
data flow recording. However, these three tasks constituted the major
part (approximately 60\%) of the CPU utilization and their data flow
constituted more than 96\% of the overall system data flow bandwidth.

\subsubsection{Robotics System-Level Control Flow Model}

In general, task activations in the robotics system are dependent
on message queues. Basically, each task is set to block on a
specific message queue until a message is received in that queue.
When a message is received, the task is activated, performs
some action, and is finally set to wait for another message to
arrive. Chains of messages and task activations, in turn, are
initiated by occurrence of external events, such as hardware
interrupts at arrival of peripheral input.

\subsubsection{Robotics Data Flow Model}\label{rdfm}

We divide the Robotics data flow model into three parts: The per-task
state preserving structures, the inter-process communication and
peripheral I/O. As for the state preserving structures, each task has
its own local data structure used to keep track of the current task
state. This structure holds information of message data, static
variables and external feedback.  Naturally, this structure alters
during execution as the state of the task changes.

The inter-process communication is handled by the use of an
IPC layer implemented on top of the message queue primitives
in VxWorks. This layer extends the functionality of the original
message queue mechanisms. Finally, peripheral I/O, such as
motion control feedback is received through a peripheral I/O
layer.

\subsection{SAAB Avionics System Model}

In addition to the Robotics system, we also performed a minor case
study in a military aircraft radar system. In short, the task of the
system is to warn the pilot of surrounding radar stations and to offer
countermeasures. This study was less extensive in that it only covered
a part of the instrumentation aspect of the Time Machine
technology. The full scale Saab Avionics radar system holds about 90
ADA tasks running on top of the Wind River VxWorks RTOS. However, in
the scope of the case study, a reduced system with 20 central tasks
was investigated.

Dataflow was limited to inter process messages. As in the robotics
case, task activations in the Avionics system are controlled by
message arrivals on certain queues.

\section{Technique Implementations}\label{ti}

To be able to facilitate replay, there were three things we needed to
achieve: First, we needed to instrument both the VxWorks real-time
kernel and the application source codes in order to be able to extract
sufficient information of the reference execution.  Second, we needed
to incorporate the replay functionality into the WindRiver integrated
development environment (IDE). We used the \emph{Tornado 2} version of
the IDE as this is the standard IDE for developing VxWorks real-time
applications. Third, we needed to add the Time Machine mechanisms used
to perform the actual replay of the system. Even though the Tornado 2
IDE features a VxWorks-level simulator, both recording and replay
execution were performed on the actual target system. In the next
sections, we will discuss these steps one by one.

\subsection{VxWorks Instrumentation}

A common denominator for the Robotics system and the Avionics system
is that they both run on top of the Wind River VxWorks RTOS. In order
to extract the exact sequence of task interleavings, it was essential
to be able to instrument the mechanisms in VxWorks that directly
influence the system-level control flow. Therefore, we instrumented
semaphore \emph{wait} and \emph{signal} operations, message queue
blocking \emph{send} and \emph{receive}, task sleep function
\emph{taskDelay} as well as preemptive scheduling decisions.

\subsubsection{Blocking System Call Instrumentation Layer}

To instrument the blocking task delay-, semaphore- and message
queue primitives mentioned above, we added an instrumentation
layer on top of the VxWorks system call API. This layer
replaces the ordinary primitives with wrappers including the
added functionality of instrumentation. Apart from this
instrumentation, the wrappers use the original primitives for
system call operation.

\subsubsection{Task Switch Hook}

With the blocking system call instrumentation layer, we are able
to monitor and log what possibly blocking system calls are
invoked. However, in order to see which of the invoked calls
actually leads to task interleavings, we need to be able to insert
probes into the scheduling mechanisms of the kernel.
Fortunately, VxWorks provides several hooks, implemented as
empty callback functions included in kernel mechanisms, such
as a TaskSwitchHook in the task switch routine. These hooks
can be used for instrumentation purposes, making it possible to
monitor and log sufficient information of each task switch in
order to be able to reproduce it.

\subsubsection{Preemption Instrumentation}

As some of the task interleavings are asynchronous, that is, their
occurrence are initiated by asynchronous events such as hardware
interrupts, they do not have an origin in the logic control flow of
task execution that blocking system calls do. To reproduce these
events and interleavings correctly in a replay execution, we need to
be able to pinpoint and monitor their exact location of occurrence. In
other words, we need some sort of \emph{unique marker} to
differentiate between different program states.  The program counter
value of the occurrence of the event is a strong candidate for such a
unique marker. However, as program counter values can be revisited in
loops, subroutines and recursive calls, additional information is
needed. To provide such data, we use the information of the task
context available from the task control block in the
TaskSwitchHook. This task context is represented in the contents of
the register set and the task stack. To avoid the massive overhead
introduced by sampling the entire contents of these areas, we use
checksums instead \cite{thane2003}. Although these checksums are not
truly unique, they strongly aid in differentiating between program
states.

Other solutions to this problem have been proposed, such as hardware
instruction counters \cite{cargill1987} and software instruction
counters \cite{mellor-crummey1989}, but these are not suitable in the
VxWorks case due to lack of kernel instrumentation possibilities,
large overheads or specialized hardware requirements.

\subsection{ABB Robotics Instrumentation}

In contrast to the system-level control flow instrumentation, which is
handled on the VxWorks RTOS level, some of the data flow
instrumentation needs to be handled on the application level. This is
due to the fact that all of the data used to represent the state of
the task execution need to be identified explicitly in the source code
and instrumented for recording. Such data include static and global
variables and structures holding information that can be altered
during the entire temporal scope of system execution.

As stated in Section~\ref{rdfm}, in the Robotics system, these data
are grouped together in static structures, individually designed for
each task. To minimize the amount of information stored in each
invocation of a task, we used filters that separated the type of data
that was prone of changes during run-time from the type of data that
was assigned values during system initialization and kept these values
throughout the execution. The latter are not recorded during
run-time. These filters were constructed from information gathered
empirically during test-runs of the system.

\begin{figure}
\centering
\includegraphics{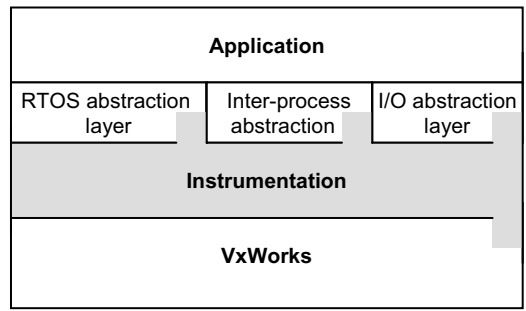}
\caption{Instrumentation layer in the system model.}
\label{instrumentation}
\end{figure}

To be able to reproduce interaction with an external context and
inter-task communication, the peripheral I/O and the inter-task
communication message queues are instrumented in two operating system
abstraction layers, similar to that described in Section 3.1.1. This
solution gives the instrumentation a quality of transparency, making
it less sensitive to changes in the application code.

However, the part of the data flow recording that is concerned with
the reproduction of state preserving structures is performed by
probing functions inserted at various locations in the application
code. A more thorough discussion on how and where these probes should
be inserted in the code is given in our previous papers
\cite{thane2003}\cite{huselius2003}. A summarizing overview of system
instrumentation can be viewed in Figure~\ref{instrumentation}. In the
figure, the gray area represents the instrumentation layer, which is
slightly integrated into different parts of the RTOS and some
application abstraction layers.

\subsection{SAAB Avionics Instrumentation}

Since both the Robotics and the Avionics system run on top of VxWorks
and the RTOS-level instrumentation is application independent, the
instrumentation of the Avionics system-level control flow was
implemented in a very similar fashion.  However, one aspect had to be
taken into account. In contrast to the Robotics system, the Avionics
system was implemented in Ada. As the Ada runtime environment is added
as a layer on top of VxWorks, this layer had to be altered in order to
be able to monitor rendezvous and other Ada synchronization
mechanisms.  This instrumentation allowed for the Ada runtime
environment to use instrumented versions of the VxWorks
synchronization system calls instead of the original versions.

As for the data flow, this study focused on inter-process
communication. Messages were logged in cyclic buffers, dimensioned on
a per-process basis, at the receiver. State preserving structures and
peripheral I/O were not considered as in the ABB Robotics
case. However, the benchmark figures of the inter-process
communication recordings give a hint of the overall overhead
incorporated in a full-scale instrumentation.

\subsection{Time Machine}

Once the code is properly instrumented, we are able to record any
execution of the system in order to facilitate replay of that very
execution. The next step is to implement the mechanisms of the time
machine that actually performs the replay. These mechanisms were
implemented in an add-on to the Tornado 2 IDE.

\subsubsection{The Historian -- Starting the Replay Execution}

\begin{figure}
\centering
\includegraphics{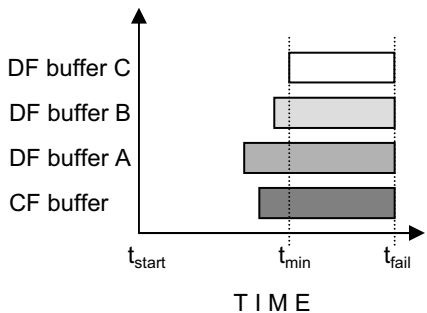}
\caption{Pruning of buffer entries. Entries to the left of $t_{min}$ are
discarded.}
\label{pruning}
\end{figure}

In our Time Machine replay system, the task of the Historian is
to analyze the data flow- and system-level control flow
recordings of a reference execution. We used basic cyclic FIFO
buffers for recording. Combined with the fact that the cyclic
buffers are of a finite length and memory resources are scarce,
this rarely leads to a situation where all recorded information is
available at the end of the reference execution. As these
recordings most often will be of a different temporal length,
some sort of pruning is needed in order to discard those entries
that are out of the consistent temporal scope of all buffers. In
other words, all tasks that are to be replayed needs information
from both the control flow recording (one per system) and data
flow recordings (one per task). Since buffers are dimensioned
using a discrete number of entries and not continuous time, we
will practically always end up in a situation where some buffers
cover a longer span of time than others. As this information is
unusable, it must be detected and discarded. This operation is
performed by the Historian as depicted in Figure~\ref{pruning}.

In addition, the Historian has the responsibility to find a consistent
state from which the replay executions can be started
\cite{huselius2003}. Again considering Figure~\ref{pruning}, if such a
starting point exists, it is located in between $t_{min}$ and
$t_{fail}$, where sufficient information of all instrumented tasks is
available. When this operation is performed, the Historian sets up the
structures in the target system used to reproduce the data flow of the
reference execution. A more elaborate description on how a starting
state is found and a replay execution is prepared is presented in a
recent paper by Huselius et. al. \cite{huselius2003}.

\subsubsection{The Time Traveler}

As the replay execution is started, the Time Traveler interacts with
the debugger and, given the information provided by the Historian and
the breakpoints visited in the program, allows recreation of the
system state for any given time in the scope of the replay execution
\cite{thane2003}.

\subsection{IDE and Target System Integration}

Tornado 2 is an integrated development environment including a text
editor, a project/workspace handler, a gcc compiler, a target
simulator and a gdb-based debugger, capable of performing
multi-threaded debugging with no requirements on deterministic
reproduction of execution orderings.

\subsubsection{Tornado 2 IDE Architecture and WTX}

Debugging in the Tornado 2 environment is performed by means of remote
debugging. That is, the debugging session is executed on-target and
controlled from the development environment via communication with an
on-target task.

\begin{figure}
\centering
\includegraphics{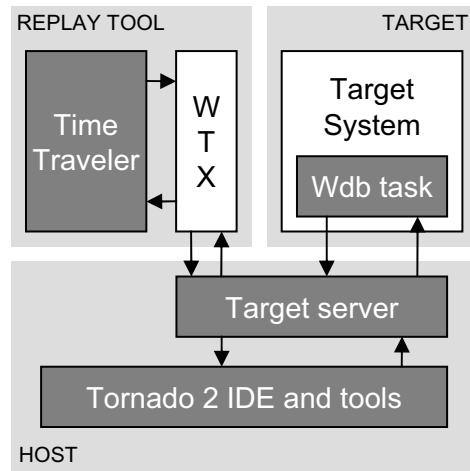}
\caption{Target system, IDE and Time Machine integration.}
\label{integration}
\end{figure}

To handle all communication with the target system, a hostbased target
server is included in the Tornado 2 IDE. All tools that seek
interaction with the target system are able to connect as clients to
the target server and issue requests of target operations. To provide
tool vendors with a possibility to create their own add-ons to the
Tornado 2 IDE, a programming interface is provided. The Wind River
Tool Exchange (WTX) API enables developers to create tools that
communicate and interact directly with the VxWorks target server. For
the implementation of our Time Machine system, the Historian and the
Time Traveler were integrated and implemented as a WTX tool, able to
connect with a running target server and to force a replay execution
upon the system running on that target. The structure of the Tornado 2
IDE, the Time Machine and the target system interactions is depicted
in Figure~\ref{integration}.

\subsubsection{Wdb Task}

To handle the on-target debugging operation, VxWorks provides a
dedicated task, the \emph{Wdb task}. This task handles requests of
breakpoints, watches, single-stepping and other debugging functions
from the IDE. These functions are used by the Time Traveler via the
WTX interface and the target server in order to control the replay
execution.

\subsubsection{Breakpoints}

Breakpoints play a central role in the interaction between the time
machine and the target system. As described by the Deterministic
Replay method \cite{thane2003}, breakpoints are set at every point of
possible task interleaving and as they are encountered in the target
system, their occurrence is echoed from the Wdb task through the WTX
and into the event handler of the Time Traveler. Based on the
individual features of each breakpoint, the state of the replay
execution can be deduced and the Time Traveler replay engine will
force the desired behavior on the target system

\subsubsection{Debugging Mode}

Debugging in VxWorks can be performed in two different modes: Task
mode and system mode. The difference is that when in system mode, an
encountered breakpoint will halt the entire system, including the
operating system. In task mode, a breakpoint will only halt the task
that encountered it, leaving all other tasks free for execution.

Ideally, since the investigated applications are pseudoparallel,
system mode debugging should be used. This would help in guaranteeing
the correct ordering of events and task interleavings in the replay
execution since no task is able to continue execution and corrupt the
system-level control flow if the entire system is halted. However, we
experienced problems when trying to reproduce this ordering in system
mode debugging regarding incapability of task suspension. This is due
to the fact that no tasks can be explicitly suspended from execution
by an external operation (such as requests made from the time traveler
tool) in system mode debugging. In addition, the locking of the Wdb
task substantially complicated communication between the target system
and the IDE, making thorough investigation of the target state more
difficult. Therefore, task mode debugging is used and the correct
ordering of events in the replay execution is explicitly forced upon
the system by means of the Time Traveler replay engine.

\section{Benchmark}\label{bm}

One of the issues of these case studies was to investigate
whether the overhead incorporated by system-level control flow-
and data flow instrumentation was acceptable in a full-scale
complex industrial real-time application or not. In order to
resolve this issue, we performed benchmarks measuring
instrumentation mechanism CPU load and memory usage in
both systems. Since the instrumentation is yet to be optimized
and the benchmarking tests are performed under worst-case
scenario conditions, many results might be rather pessimistic.

\subsection{ABB Robotics}

\begin{figure}
\centering
\includegraphics{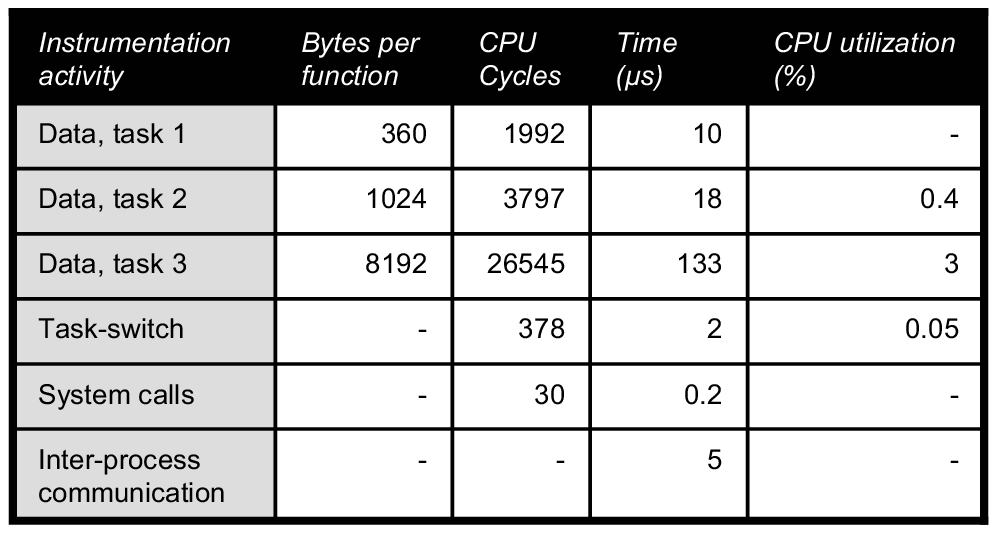}
\caption{Benchmarking resulsts from the Robotics study.}
\label{ABBbenchmark}
\end{figure}

In the ABB robotics case, we timestamped the entry and the exit of all
instrumentation and recording mechanisms. This gave us a possibility
of extracting the execution time of the software probes. In addition,
we measured the frequency and recording size of the data flow
instrumentation mechanisms. The achieved results are presented in
Figure~\ref{ABBbenchmark}. The \emph{bytes per function} column shows
the number of bytes logged by each iteration of an instrumentation
function, and as each task instance has one and only one data flow
instrumentation function, this figure also represents the number of
bytes stored in each task instance of a task. The \emph{CPU cycles}
and the \emph{Time} columns present the execution time spent in each
instrumentation function and the \emph{CPU utilization} column shows
the percentage of all CPU time spent in each instrumentation
function. Where results are left out, these are discarded due to their
insignificant interference with memory or CPU utilization. We note
that task 3 has a combination of a high frequency and large state
preserving structures, resulting in the largest monitoring overhead
(3\% CPU utilization).

\subsection{SAAB Avionics}

\begin{figure}
\centering
\includegraphics{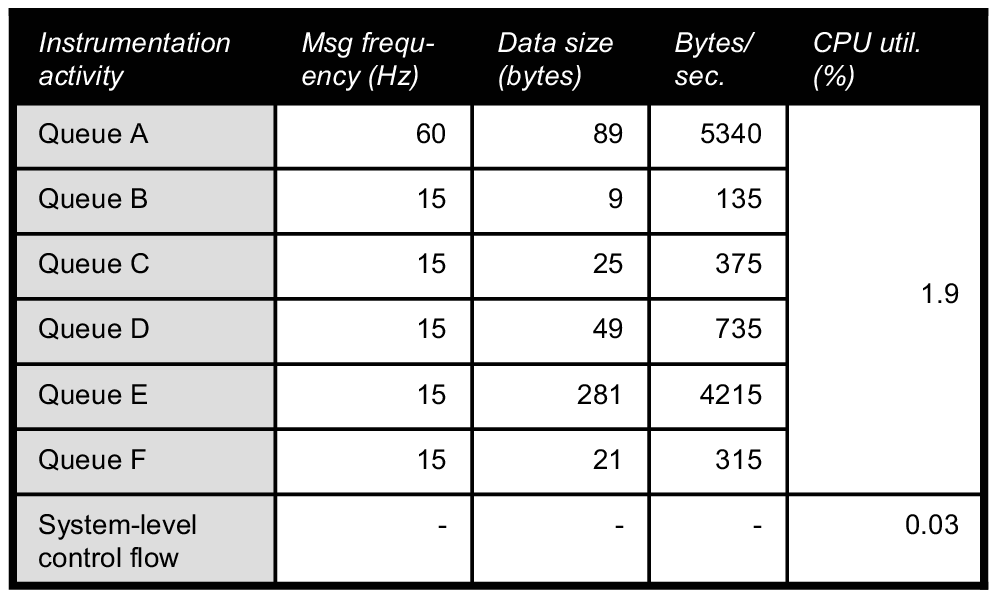}
\caption{Avionics system benchmarking results.}
\label{SAABbenchmark}
\end{figure}

As the data instrumentation in the Avionics system is performed solely
on message queues, the data flow benchmark is made in a per-queue
fashion. Since there are major differences in message arrival
frequencies and message size between the different queues, only the
six most memory-consuming message queues are presented in the
benchmark results. Out of the 17 instrumented message queues, these
six consume 99\% of all message queue memory bandwidth. The results of
the Avionics benchmarking study are shown in
Figure~\ref{SAABbenchmark}, presented in the form of memory
utilization for logging (in the bytes/second column) and CPU
utilization (in the rightmost column). As in the robotics case, it is
the combination of a high frequency and a large size of data (such as
the one of Queue A and Queue E) that has the most significant
implications on the instrumentation memory utilization.

\section{Conclusions}\label{cs}

With this paper, we have shown that complex real-time system debugging
is feasible using the \emph{Deterministic Replay} technique and the
\emph{Time Machine} tool. This is true not only for specialized
academic systems, but also for full-scale industrial real-time
systems. Furthermore, we have shown that it is possible to achieve a
high level of transparency and portability of the method by placing
much of the instrumentation in system call-, inter-process
communication- and peripheral I/O layers, rather than in the
application source code.  Both case studies presented here indicate a
small CPU utilization overhead of 0.03--0.05\% for system-level
control flow instrumentation. The data flow instrumentation has proven
more temporally substantial, but has stayed in the fully acceptable
interval of 1.9--3.0\%. As for memory utilization, the ABB Robotics
instrumentation required a bandwidth of 2 MB/s and the Saab Avionics
system called for approximately 12-15 kB/s for both system-level
control flow and data flow logging.  Looking at the size of these
systems and the resources available, such a load is definitely
affordable.

\section{Future Work}\label{fw}

We have successfully applied the time machine approach in a number of
applications running on different operating systems, hardware and
debuggers \cite{thane2003}. However, we have learned that it is
necessary to carefully analyze the target system's data flow with
respect to what data is re-executed, re-transmitted and what data has
external origin in order not to forego something that may inhibit
deterministic re-execution or that we do not record to much. Missing
out on information is a serious problem and is something we have begun
to address. We have started to look into how to automatically derive
tight sets of possible data derived from what we actually have
recorded. Another issue that needs to be considered is replay of
applications which use vast amounts of information, e.g., real-time
database applications. In such applications the ``state preserving
variables'' are very substantial and require some kind of incremental
snapshot algorithm to be manageable.

\bibliography{../../Bibliography/bib}

\end{document}